\RequirePackage{snapshot}
\documentclass[conference]{IEEEtran}

\usepackage{amsfonts,amsmath,amssymb,mathtools}
\usepackage{flushend}
\usepackage{bigints}
\usepackage{hhline}
\usepackage{marvosym} 

\usepackage{tgpagella}
\usepackage[T1]{fontenc}

\usepackage{graphicx}
\usepackage{booktabs, multicol, multirow}
\usepackage{indentfirst}
\usepackage{subfigure}
\usepackage[font=footnotesize]{caption}
\DeclareCaptionLabelFormat{andtable}{#1~#2  \&  \tablename~\thetable}
\DeclareCaptionLabelFormat{andfigure}{\tablename~\thetable~\&~#1~#2}

\usepackage{marginnote}
\usepackage[usenames,dvipsnames]{color}
\usepackage[table]{xcolor}    
\definecolor{fullred}{rgb}{0.85,.0,.1} 
\definecolor{navyblue}{rgb}{.0,.0,.5}
\definecolor{bleudefrance}{rgb}{0.19, 0.55, 0.91}
\definecolor{bluegray}{rgb}{0.18, 0.36, 0.6}

\usepackage[textsize=tiny]{todonotes}

\usepackage{array}
\newcolumntype{+}{>{\global\let\currentrowstyle\relax}}
\newcolumntype{^}{>{\currentrowstyle}}

\newcolumntype{C}[1]{>{\centering\arraybackslash}p{#1}}

\usepackage[numbers,sort]{natbib}
\usepackage[bookmarks=true, colorlinks=true, urlcolor=Blue]{hyperref}
\usepackage{multicol}

\newcommand{\ME}{Sudeep Pillai}
\newcommand{\PAPERAUTHORS}{Sudeep Pillai and John J. Leonard}
\newcommand{\PAPERTITLE}{Monocular SLAM Supported Object Recognition}
\newcommand{\PAPERTITLEFMT}{Monocular SLAM Supported\\Object Recognition}
\newcommand{\PAPERKEYWORDS}{Object Recognition; Computer Vision; Perception}

\hypersetup{pdfauthor={\ME},%
            pdftitle={\PAPERTITLE},%
            pdfsubject={\PAPERTITLE},%
            pdfkeywords={\PAPERKEYWORDS},%
            pdfproducer={LaTeX},%
            pdfcreator={pdfLaTeX}
}

\begin{document}

\title{\PAPERTITLEFMT{}\vspace{-0.5ex}}


\author{\authorblockN{\PAPERAUTHORS{}}
\authorblockA{
Computer Science and Artificial Intelligence Laboratory\\
Massachusetts Institute of Technology\\
\{\href{mailto:spillai@csail.mit.edu}{spillai}, \href{mailto:jleonard@csail.mit.edu}{jleonard}\}@csail.mit.edu}}

\maketitle

\vspace{-1ex}
\begin{abstract}
In this work, we develop a monocular SLAM-aware object recognition system that
is able to achieve considerably stronger recognition performance, as compared to
classical object recognition systems that function on a frame-by-frame basis. By
incorporating several key ideas including multi-view object proposals and
efficient feature encoding methods, our proposed system is able to detect and
robustly recognize objects in its environment using a single RGB camera in
near-constant time. Through experiments, we illustrate the utility of using such
a system to effectively detect and recognize objects, incorporating multiple
object viewpoint detections into a unified prediction hypothesis. The
performance of the proposed recognition system is evaluated on the UW RGB-D
Dataset, showing strong recognition performance and scalable run-time
performance compared to current state-of-the-art recognition systems.

\end{abstract}

\IEEEpeerreviewmaketitle

\section{Introduction}
\label{sec-introduction} Object recognition is a vital component in a robot's
repertoire of skills. Traditional object recognition methods have focused on
improving recognition performance (Precision-Recall, or mean Average-Precision)
on specific datasets~\cite{everingham2010pascal, ILSVRC15}. While these datasets
provide sufficient variability in object categories and instances, the training
data mostly consists of images of arbitrarily picked scenes and/or
objects. Robots, on the other hand, perceive their environment as a continuous
image stream, observing the same object several times, and from multiple
viewpoints, as it constantly moves around in its immediate environment. As a
result, object detection and recognition can be further bolstered if the
robot were capable of simultaneously localizing itself and mapping (SLAM) its
immediate environment - by integrating object detection evidences
across multiple views.




%
\begin{figure}[!t]
  \centering
  \begin{tabular}{c}
    \includegraphics[width=\columnwidth]{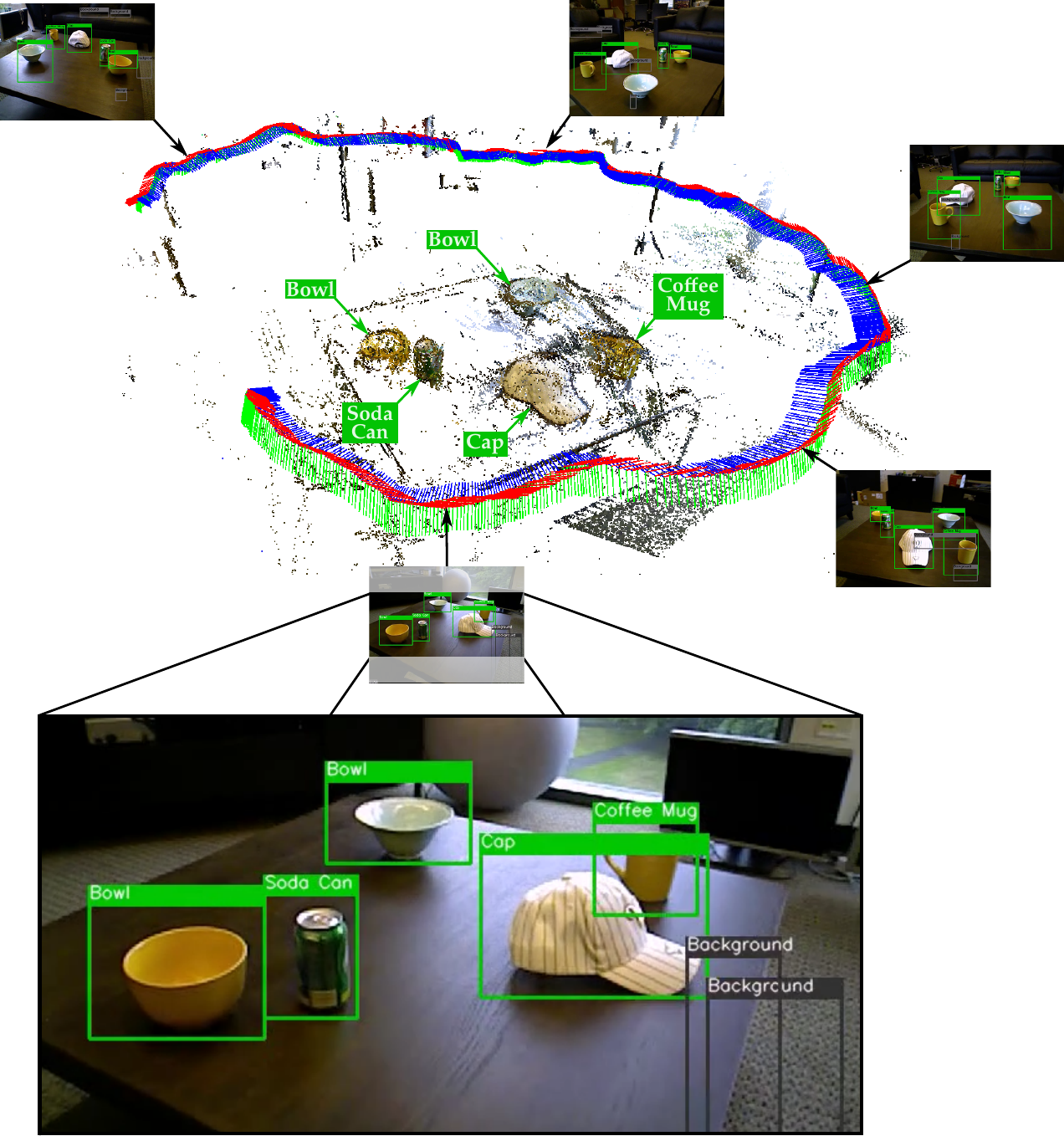}
  \end{tabular}
  \caption{The proposed SLAM-aware object recognition system is able to robustly
    localize and recognize several objects in the scene, aggregating detection evidence across multiple views. Annotations in white are provided for clarity and are actual predictions
    proposed by our system.}
  \label{fig:intro-fig}
\end{figure}


We refer to a ``SLAM-aware'' system as - one that has access to the map of its
observable surroundings as it builds it incrementally and the location of its
camera at any point in time. This is in contrast to classical recognition
systems that are ``SLAM-oblivious'' - those that detect and recognize objects on
a frame-by-frame basis without being cognizant of the map of its environment, the
location of its camera, or that objects may be situated within these
maps. \textit{In this paper, we develop the ability for a SLAM-aware system to
robustly recognize objects in its environment, using an RGB camera as its only
sensory input} (Figure~\ref{fig:intro-fig}).

We make the following contributions towards this end: Using state-of-the-art
semi-dense map reconstruction techniques in monocular visual SLAM as
pre-processed input, we introduce the capability to propose multi-view
consistent object candidates, as the camera observes instances of objects across
several disparate viewpoints. Leveraging this object proposal method, we
incorporate some of the recent advancements in bag-of-visual-words-based (BoVW)
object
classification~\cite{jegou2010aggregating,delhumeau2013revisiting,arandjelovic2013all}
and efficient box-encoding methods~\cite{van2014fisher} to enable strong
recognition performance.  The integration of this system with a monocular
visual-SLAM (vSLAM) back-end also enables us to take advantage of both the
reconstructed map and camera location to significantly bolster recognition
performance. Additionally, our system design allows the run-time performance to
be scalable to a larger number of object categories, with near-constant run-time
for most practical object recognition tasks.


%
\begin{figure*}[!t]
    \centering
    \includegraphics[width=2\columnwidth]{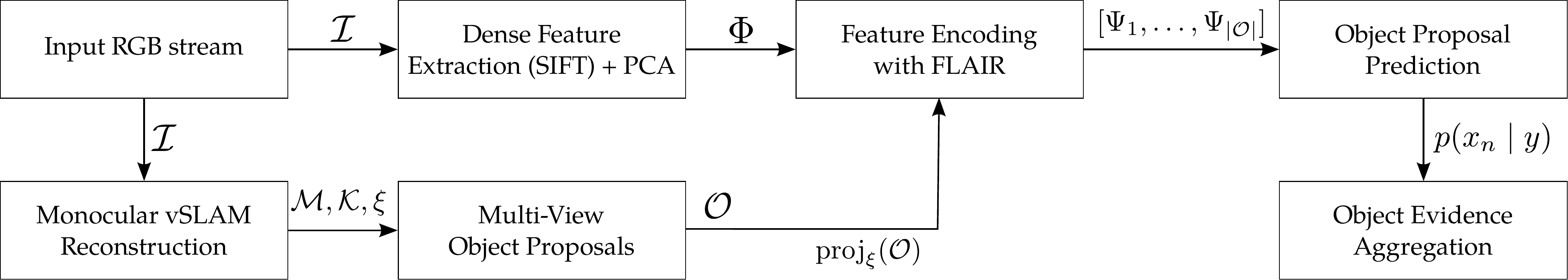}
    \caption{Outline of the SLAM-aware object recognition pipeline. Given an
input RGB image stream $\mathcal{I}$, we first reconstruct the scene in a
semi-dense fashion using an existing monocular visual-SLAM implementation
(ORB-SLAM) with a semi-dense depth estimator, and subsequently extract relevant
map $\mathcal{M}$, keyframe $\mathcal{K}$ and pose information $\xi$. We perform
multi-scale density-based segmentation on the reconstructed scene to obtain
object proposals $\mathcal{O}$ that are consistent across multiple views. On
each of the images in the input RGB image stream $\mathcal{I}$, we compute
Dense-SIFT ($\mathbb{R}^{128}$) + RGB ($\mathbb{R}^{3}$) and reduce it to $\Phi
\in \mathbb{R}^{80}$ via PCA. The features $\Phi$ are then used to efficiently
encode each of the projected object proposals $\mathcal{O}$ (bounding boxes of
proposals projected on to each of the images with known poses $\xi$) using VLAD
with FLAIR, to obtain $\Psi$. The resulting feature vector $\Psi$ is used to
train and predict likelihood of target label/category $p(x_i\mid y)$ of the
object contained in each of the object proposals. The likelihoods for each
object $o \in \mathcal{O}$ are aggregated across each of the viewpoints $\xi$ to
obtain robust object category prediction.}
    \label{fig:recognition-pipeline}
\end{figure*}

We present several experimental results validating the improved object
proposition and recognition performance of our proposed system: (i) The
system is compared against the current
state-of-the-art~\cite{lai2012detection,laiunsupervised} on the UW-RGBD
Scene~\cite{lai2011large,laiunsupervised} Dataset. We compare the improved
recognition performance of being SLAM-aware, to being SLAM-oblivious (ii) The
multi-view object proposal method introduced is shown to outperform single-view
object proposal strategies such as BING~\cite{cheng2014bing} on the UW-RGBD
dataset, that provide object candidates solely on a single-view. (iii) The
run-time performance of our system is analysed, with specific discussion on the
scalability of our approach, compared to existing state-of-the-art
methods~\cite{lai2012detection, laiunsupervised}.


%



\section{Related Work}
\label{sec:related-work}


We discuss some of the recent developments in object 
proposals, recognition, 
and semi-dense monocular visual SLAM literature that has sparked the ideas explained in this paper.

%
%
\textbf{Sliding window techniques and DPM} In traditional state-of-the-art
object detection, HOG~\cite{dalal2005histograms} and
deformable-part-based-models (DPM) proposed by~\citet{felzenszwalb2010object}
have become the norm due to their success in recognition performance. These
methods explicitly model the shape of each object and its parts via
oriented-edge templates, across several scales. Despite its reduced
dimensionality, the template model is scanned over the
entire image in a sliding-window fashion across multiple scales for each object
type that needs to be identified. This is a highly limiting
factor in scalability, as the run-time performance of the system is directly
dependent on the number of categories identifiable. While techniques have been
proposed to scale such schemes to larger object categories~\cite{dean2013fast},
they incur a drop in recognition performance to trade-off for speed.


\textbf{Dense sampling and feature encoding methods} Recently, many of the
state-of-the-art techniques~\cite{lazebnik2006beyond,van2014fisher} for generic
object classification have resorted to dense feature extraction. Features are
densely sampled on an image grid~\cite{bosch2007image}, described, encoded and
aggregated over the image or a region to provide a rich description of the
object contained in it. The aggregated feature encodings lie as feature vectors
in high-dimensional space, on which linear or kernel-based classification
methods perform remarkably well. Among the most popular encoding schemes include
Bag-of-Visual-Words (BoVW)~\cite{csurka2004visual,sivic2003video}, and more
recently Super-Vectors~\cite{zhou2010image}, VLAD~\cite{jegou2010aggregating},
and Fisher Vectors~\cite{perronnin2010improving}. In the case of BoVW, a
histogram of occurrences of codes are built using a vocabulary of finite size $V
\in \mathbb{R}^{K \times D}$. VLAD and Fisher Vectors, in contrast, aggregate
residuals using the vocabulary to estimate the first and second order moment
statistics in an attempt to reduce the loss of information introduced in the
vector-quantization (VQ) step in BoVW. Both VLAD and Fisher Vectors have been
shown to outperform traditional BoVW
approaches~\cite{jegou2010aggregating,perronnin2010improving,chatfield2011devil},
and are used as a drop-in replacement to BoVW; we do the same utilizing VLAD as
it provides a good trade-off between descriptiveness and computation time.


\textbf{Object Proposals} Recently, many of the
state-of-the-art techniques in large-scale object recognition systems have
argued the need for a category-independent object proposal method that provides candidate
regions in images that may likely contain objects. Variants of these include
Constrained-Parametric Min-cuts (CPMC)~\cite{carreira2010constrained}, Selective
Search~\cite{uijlings2013selective}, Edge Boxes~\cite{zitnick2014edge},
Binarized Normed Gradients (BING)~\cite{cheng2014bing}. The object candidates
proposed are category-independent, and achieve detection rates (DR) of 95-99\% at
0.7 intersection-over-union~(IoU\footnote{Intersection-over-Union (IoU) is a
common technique to evaluate the quality of candidate object proposals with
respect to ground truth. The intersection area of the ground truth bounding box
and that of the candidate is divided by the union of their areas.}) threshold, by generating about 1000-5000
candidate proposal windows~\cite{hosang2014good,zitnick2014edge}. This
dramatically reduces the search space for existing sliding-window approaches
that scan templates over the entire image, and across multiple scales; however, it
still bodes a challenge to accurately classify irrelevant proposal windows as
background.  For a thorough evaluation of the state-of-the-art object proposal
methods, and their performance, we refer the reader to~\citet{hosang2014good}.


\textbf{Scalable Encoding with Object Proposals} As previously addressed,
sliding-window techniques inherently deal with the scalability issue, despite
recent schemes to speed-up such an approach. BoVW, on the contrary, handle this
scalability issue rather nicely since the histograms do not particularly encode
spatial relations as strongly. This however, makes BoVW approaches lack the
ability to localize objects in an image. The advent of category-independent
object proposal methods have subsequently opened the door to bag-of-words-driven
architectures, where object proposal windows can now be described via existing
feature encoding methods. Most recently,~\citet{van2014fisher} employ a novel
box-encoding technique using integral histograms to describe object proposal
windows with a run-time independent of the window size of object proposals
supplied. They report results with an 18x speedup over brute-force BoVW encoding
(for 30,000 object proposals), enabling a new state-of-the-art on the
challenging 2010 PASCAL VOC detection task. Additionally their
proposed system ranks number one in the official ImageNet 2013 detection
challenge, making it a promising solution to consider for robotics applications.




\textbf{Multi-view Object Detection} While classical object detection methods
focus on single-view-based recognition performance, some of these methods have
been extended to the multi-view
case~\cite{thomas2006towards,collet2010efficient}, by aggregating object
evidence across disparate views. \citet{lai2012detection} proposed a
multi-view-based approach for detecting and labeling objects in a 3D environment
reconstructed using an RGB-D sensor. They utilize the popular HOG-based
sliding-window detectors trained from object views in the RGB-D
dataset~\cite{lai2011large,laiunsupervised} to assign class probabilities to
pixels in each of the frames of the RGB-D stream. Given co-registered image and
depth, these probabilities are assigned to voxels in a discretized reconstructed
3D scene, and further smoothed using a Markov Random Field (MRF). Bao et
al.~\cite{bao2011semantic,bao2012semantic} proposed one of the first approaches
to jointly estimate camera parameters, scene points and object labels using both
geometric and semantic attributes in the scene. In their work, the authors
demonstrate the improved object recognition performance, and robustness by
estimating the object semantics and SfM jointly. However, the run-time of ~20
minutes per image-pair, and the limited object categories identifiable makes the
approach impractical for on-line robot operation. Other works
~\cite{salas2013slam++,castle2010combining,civera2011towards,bo2011hierarchical,guptaECCV14}
have also investigated object-based SLAM, SLAM-aware, and 3D object recognition
architectures, however they have a few of glaring concerns: either (i) the
system cannot scale beyond a finite set of object instances (generally limited
to less than 10), or (ii) they require RGB-D input to support both detection and
pose estimation, or (iii) they require rich object information such as 3D models
in its database to match against object instances in a brute-force manner.

\begin{figure*}[!th]
  \centering
  \includegraphics[width=2\columnwidth]{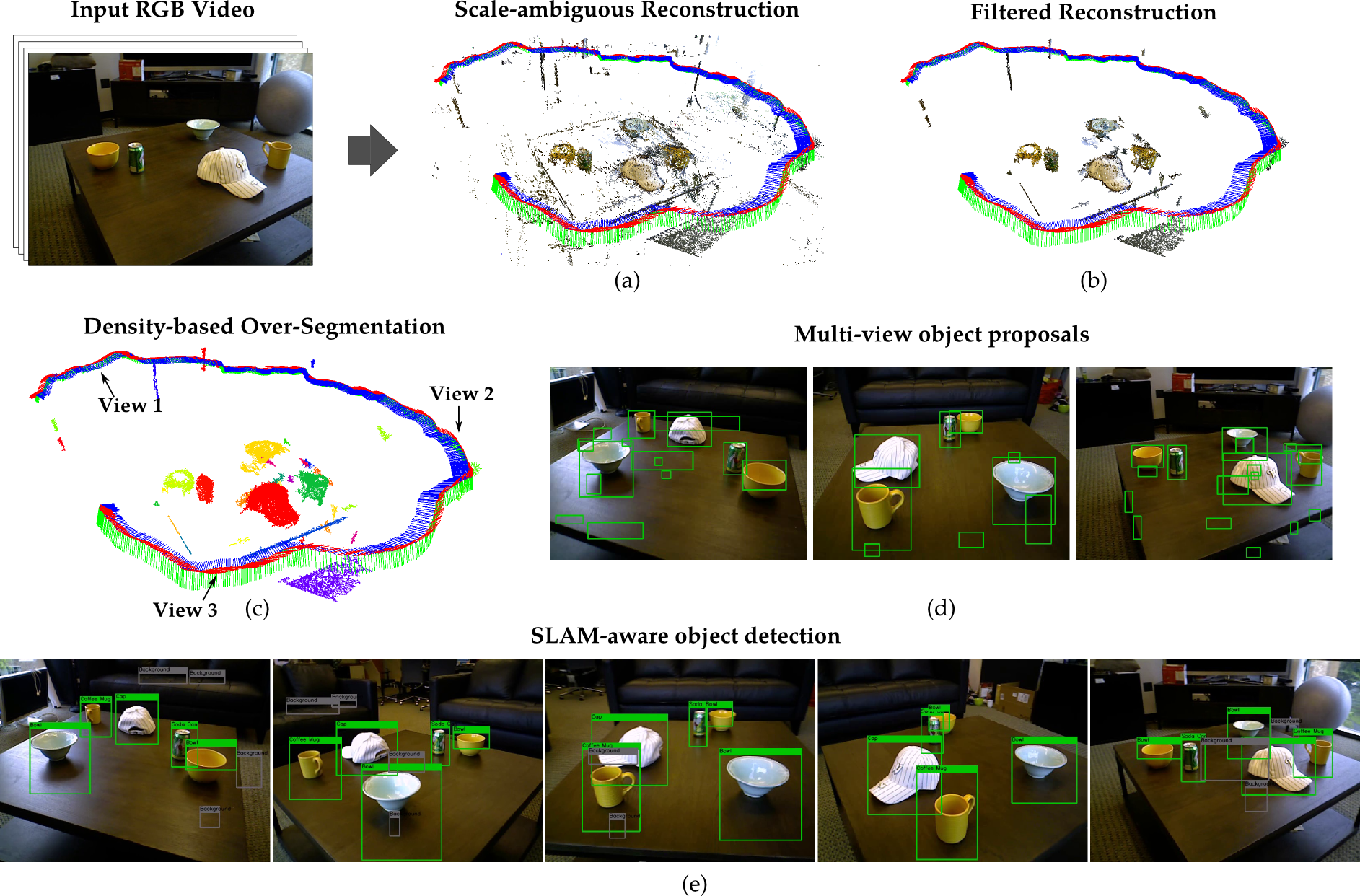}
  \caption{An illustration of the multi-view object proposal method and
    subsequent SLAM-aware object recognition. Given an input RGB image stream, a
    scale-ambiguous semi-dense map is reconstructed (a) via the 
    ORB-SLAM-based~\cite{mur2015orb} semi-dense mapping solution. The reconstruction
    retains edges that are consistent across multiple views, and is employed in
    proposing objects directly from the reconstructed space. The resulting
    reconstruction is (b) filtered and (c) partitioned into several segments using a
    multi-scale density-based clustering approach that teases apart objects (while
    filtering out low-density regions) via the semi-dense edge-map
    reconstruction. Each of the clustered regions are then (d) projected on to each
    of individual frames in the original RGB image stream, and a bounded candidate
    region is proposed for subsequent feature description, encoding and
    classification. (e) The probabilities for each of the proposals per-frame are
    aggregated across multiple views to infer the most likely object label.}
  \label{fig:multi-view-objectness}
\end{figure*}

\section{Monocular SLAM Supported \\Object Recognition}
\label{sec:proc-procedure}
This section introduces the algorithmic components of our method.
We refer the reader to Figure~\ref{fig:recognition-pipeline} that illustrates the
steps involved, and provide a brief overview of our system. 

\subsection{Multi-view Object Proposals}
\label{sec:proc-objectness} 

Most object proposal strategies use either superpixel-based or edge-based
representations to identify candidate proposal windows in a single image that
may likely contain objects. Contrary to classical per-frame object proposal
methodologies, robots observe the same instances of objects in its environment
several times and from disparate viewpoints. It is natural to think of object
proposals from a spatio-temporal or reconstructed 3D context, and a key
realization is the added robustness that the temporal component provides in
rejecting spatially inconsistent edge observations or candidate proposal
regions. Recently,~\citet{engel2014lsd} proposed a scale-drift aware monocular
visual SLAM solution called LSD-SLAM, where the scenes are reconstructed in a
semi-dense fashion, by fusing spatio-temporally consistent scene edges. Despite
being scale-ambivalent, the multi-view reconstructions can be especially
advantageous in teasing apart objects in the near-field versus those in the
far-field regions, and thus subsequently be useful in identifying candidate
object windows for a particular view. We build on top of
an existing monocular SLAM solution (ORB-SLAM~\cite{mur2015orb}) and augment a
semi-dense depth filtering component derived from~\cite{forster2014svo}. The
resulting reconstruction qualitatively is similar to that produced by
LSD-SLAM~\cite{engel2014lsd}, and is used for subsequent object proposal
generation. We avoided the use of LSD-SLAM as it occasionally failed over
tracking wide-baseline motions inherent in the benchmark dataset we used. 



In order to retrieve object candidates that are spatio-temporally consistent, we
first perform a density-based partitioning on the scale-ambiguous reconstruction
using both spatial and edge color information. This is done repeatedly for 4
different density threshold values (each varied by a factor of 2), producing an
over-segmentation of points in the reconstructed scene that are used as seeds
for multi-view object candidate proposal. The spatial density segmentations
eliminate any spurious points or edges in the scene, and the resulting point
cloud is sufficient for object proposals. These object over-segmentation seeds
are subsequently projected onto each of the camera views, and serve as seeds to
for further occlusion handling, refinement and candidate object proposal
generation. We cull out (i) small candidates whose window size is less than
20x20 px, (ii) occluding candidates by estimating their median depth from the
reconstruction, to avoid mis-identification and (iii) overlapping candidates
with an IoU threshold of 0.5, to avoid redundant proposals. The filtered set of
windows are subsequently considered as candidates for the classification process
downstream. Figure~\ref{fig:multi-view-objectness} illustrates the different
steps described in this section.



\subsection{State-of-the-art Bag-of-Visual-Words with Object Proposals}
\label{subsec:proc-vlad-flair}

Given the object proposals computed using the reconstructed scale-ambiguous map,
we now direct our attention to describing these proposal regions. 


%
%
\begin{figure*}[!t]
    \centering
    \includegraphics[width=2\columnwidth]{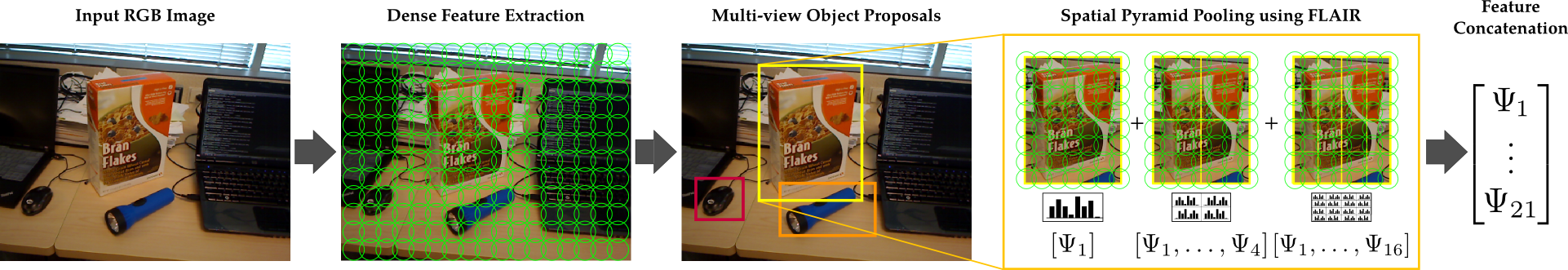}
    \caption{Various steps involved in the feature extraction
procedure. Features that are densely sampled from the image are subsequently
used to describe the multi-view object proposals using FLAIR. Each proposal is
described with multiple ([1x1], [2x2], [4x4]) spatial levels/bins via quick
table lookups in the integral VLAD histograms (through FLAIR). The resulting
histogram $\Psi$ (after concatenation) is used to describe
the object contained in the bounding box. Figure is best viewed in electronic form.}
    \label{fig:feature-extraction}
\end{figure*}

\textbf{Dense BoVW with VLAD} Given an input image
and candidate object proposals, we first densely sample the image, describing
each of the samples with SIFT + RGB color values, $\Phi_{SIFT+RGB} \in
\mathbb{R}^{131}$ i.e. Dense SIFT~(128-D) + RGB(3-D). Features are extracted with a
step size of 4 pixels, and at 4 different pyramid scales with a pyramid scale
factor of $\sqrt[]{2}$. The resulting description is then reduced to a
80-dimensional vector via PCA, called PCA-SIFT $\Phi \in \mathbb{R}^{80}$. A
vocabulary $V \in \mathbb{R}^{K\times80}$ of size $K=64$ is created via
$k$-means, using the descriptions extracted from a shuffled subset of the training data,
as done in classical bag-of-visual-words approaches. In classical BoVW, this
vocabulary can be used to encode each of the original SIFT+RGB descriptions in
an image into a histogram of occurrences of codewords, which in turn provides a
compact description of the original image. Recently, however, more
descriptive encodings such as VLAD~\cite{jegou2010aggregating} and Fisher
Vectors~\cite{perronnin2010improving} have been shown to outperform classical
BoVW
approaches~\cite{jegou2010aggregating,perronnin2010improving,chatfield2011devil}. Consequently,
we chose to describe the features using VLAD as it provides equally as strong performance
with slightly reduced computation time as compared to Fisher Vectors. 

For each of the bounding boxes, the un-normalized VLAD $\Psi \in
\mathbb{R}^{KD}$ description is computed
by aggregating the residuals of each of the descriptions $\Phi$ (enclosed within the
bounding box) from their vector-quantized centers in the vocabulary, thereby
determining its first order moment (Eq.~\ref{eqn:vlad-first-order}). \vspace{-1mm}
\begin{align}
v_k = \sum_{x_i:NN(x_i)=\mu_k} x_i - \mu_k
\label{eqn:vlad-first-order}
\end{align} 
The
description is then normalized using signed-square-rooting (SSR) or commonly
known as power normalization (Eq.~\ref{eq:signed-square-rooting}) with
$\alpha=0.5$, followed by L2 normalization, for improved recognition performance
as noted in~\cite{arandjelovic2013all}.
\begin{align} f(z) = sign(z)\vert z \vert^\alpha \quad \text{where} \quad 0 \leq
\alpha \leq 1
\label{eq:signed-square-rooting}
\end{align} Additional descriptions for each bounding region are constructed for
3 different spatial bin levels or subdivisions as noted in
~\cite{lazebnik2006beyond} (1x1, 2x2 and 4x4, 21 total subdivisions $S$), and
stacked together to obtain the feature vector $\Psi = \begin{bmatrix}
\hdots \textbf{v}_s \hdots \end{bmatrix} \in 
\mathbb{R}^{KDS}$ that appropriately describes the
specific object contained within the candidate object proposal/bounding box.\vspace{2mm}

\textbf{Efficient Feature Encoding with FLAIR} While it may be practical to
describe a few object proposals in the scene with these encoding methods, it can
be highly impractical to do so as the number of object proposals grows. To this
end, \citet{van2014fisher} introduced FLAIR - an encoding mechanism that
utilizes summed-area tables of histograms to enable fast descriptions for
arbitrarily many boxes in the image. By constructing integral histograms for
each code in the codebook, the histograms or descriptions for an arbitrary
number of boxes $B$ can be computed independent of their area. As shown
in~\cite{van2014fisher}, these descriptions can also be extended to the VLAD
encoding technique. Additionally, FLAIR affords performing spatial pyramid
binning rather naturally, with only requiring a few additional table look-ups,
while being independent of the area of $B$. We refer the reader to
Figure~\ref{fig:feature-extraction} for an illustration of the steps involved in
describing these candidate object proposals.\vspace{2mm}

\textbf{Multi-class histogram classification} Given training examples, $(x_1, y_1),\ldots,(x_n, y_n)$ where $x_i\in\mathbb{R}^{KDS}$ are the VLAD descriptions 
and $y_i\in\{1,\ldots, \mathcal{C}\}$ are the ground truth target labels,
we train a linear classifier using Stochastic Gradient Descent (SGD), given by:
\begin{align} 
E(w) = \frac{1}{n}\sum_{i=1}^{n} L(y_i, f(x_i)) + \alpha R(w)
\label{eq:sgd-classifier}
\end{align} \vspace{-1mm} where $L(y_i, f(x_i)) =
\text{log}\Big(1+\text{exp}(-y_i\textbf{w}^T\textbf{x}_i)\Big)$ is the logistic
loss function, $R(w) = \frac{1}{2} \sum_{i=1}^{n} \textbf{w}^T\textbf{w}$ is the L2-regularization term that penalizes model complexity, and $\alpha
> 0$ is a non-negative hyperparameter that adjusts the L2 regularization. A
one-versus-all strategy is taken to extend the classifiers to multi-class
categorization. For hard-negative mining, we follow~\cite{van2014fisher}
closely, bootstrapping additional examples from wrongly classified negatives for
2 hard-negative mining epochs.

%
\begin{figure*}[!tp]
    \centering
    \includegraphics[width=2\columnwidth]{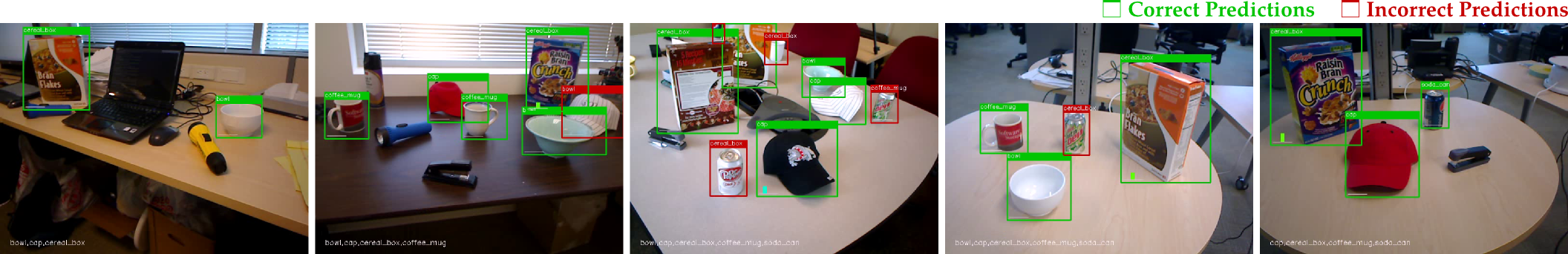}
    \caption{Illustration of \textbf{per-frame} detection results provided by our object
      recognition system that is \textit{intentionally SLAM-oblivious} (for
      comparison purposes only). Object recognition evidence is
      not aggregated across all frames, and detections are performed on a
      frame-by-frame basis. Only detections having corresponding ground truth
      labels are shown. Figure is best viewed in electronic form. }
    \label{fig:frame-detection-results}
\end{figure*}
%

%
\begin{figure*}[t]
    \centering
    \includegraphics[width=2\columnwidth]{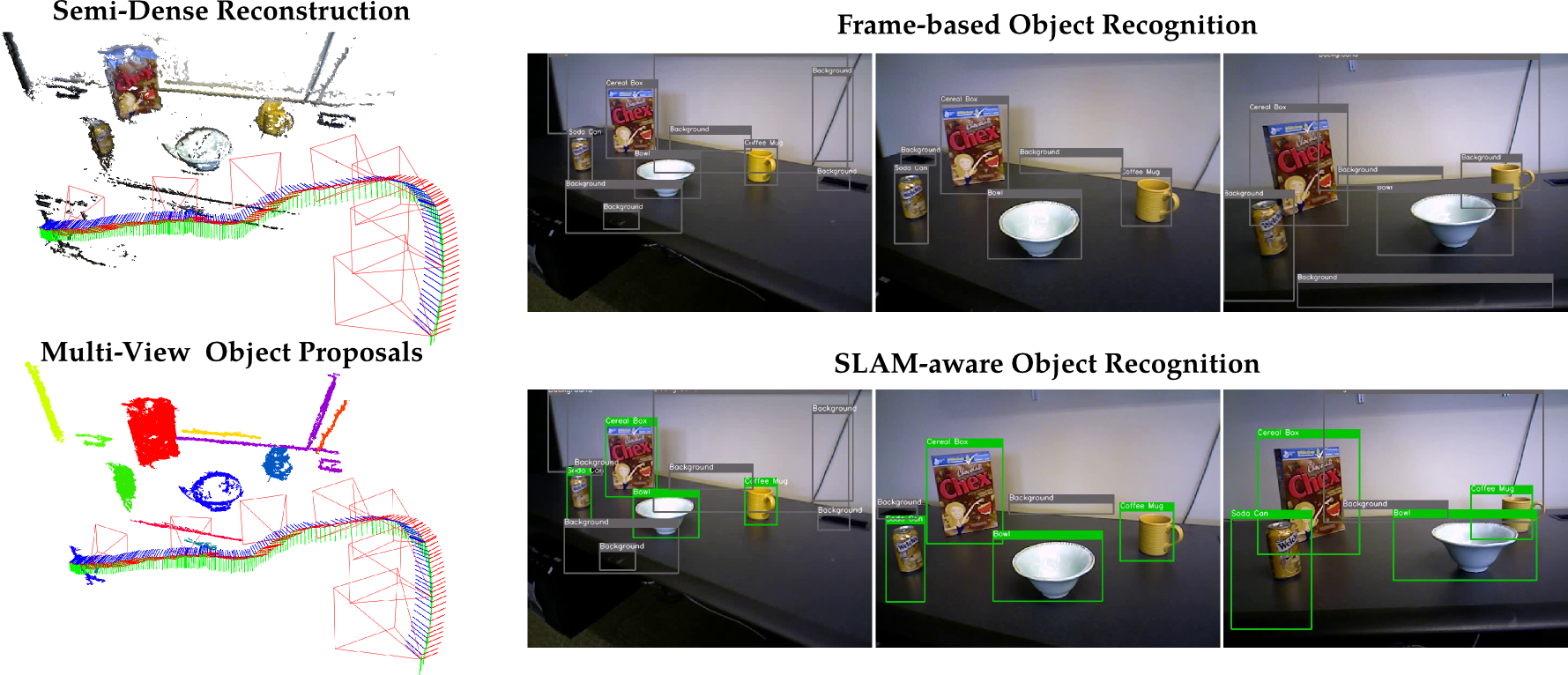}
    \caption{Illustration of the recognition capabilities of our proposed
      SLAM-aware object recognition system. Each of the object categories are detected
      every frame, and their evidence is aggregated across the entire sequence through
      the set of object hypothesis. In frame-based object recognition, predictions are
      made on an individual image basis (shown in gray).  In SLAM-aware recognition,
      the predictions are aggregated across all frames in the image sequence to
      provide robust recognition performance. The green boxes indicate correctly
      classified object labels, and the gray boxes indicate background object
      labels. Figure is best viewed in electronic form.}
    \label{fig:slam-detection-results-1}
\end{figure*}

\subsection{Multi-view Object Recognition}
\label{sec:proc-multiview-recognition} We start with the ORB-SLAM-based
semi-dense mapping solution, that feeds a continuous image stream, in order to
recover a scale-ambiguous map~$\mathcal{M}$, keyframes~$\mathcal{K}$, and poses
${\xi}$ corresponding to each of the frames in the input image stream. The
resulting scale-ambiguous reconstruction provides a strong indicator of object
presence in the environment, that we use to over-segment into several object
seeds $o \in \{1, \dots, \mathcal{O}\}$.  These object seeds are projected back
in to each of the individual frames using the known projection matrix, derived
from its corresponding viewpoint $\xi_i$. The median depth estimates of each of
the seeds are estimated in order to appropriately project non-occluding object
proposals back in to corresponding viewpoint, using a depth buffer. Using these
as candidate object proposals, we evaluate our detector on each of the
$\mathcal{O}$ object clusters, per image, providing probability estimates of
belonging to one of the $\mathcal{C}$ object classes or categories. Thus, the
maximum-likelihood estimate of the object $o \in \mathcal{O}$ can be formalized
as maximizing the data-likelihood term for all observable viewpoints (assuming
uniform prior across the $\mathcal{C}$ classes):
\begin{align}
\hat{y}^{MLE} = \underset{y\in\{1,\dots,\vert\mathcal{C}\vert\}}{\mathrm{argmax}}~p(\mathcal{D}_{o}\mid y) \quad \forall~o \in \mathcal{O}
\label{eq:multi-view-evidence}
\end{align}
where $y\in\{1,\dots,\vert\mathcal{C}\vert\}$ are
the class labels, $\mathcal{D}_{o} = \{x_{1}, \dots, x_{N}\}_{o}$ is the data observed of the
object cluster $o \in \mathcal{O}$ across $N$ observable viewpoints. In our
case, $\mathcal{D}_{o}$ refers to the bounding box of
the $o^{th}$ cluster, projected onto each of the $N$
observable viewpoints. Assuming the individual features in $\mathcal{D}_{o}$ are conditionally
independent given the class label $y$, the maximum-likelihood estimate (MLE)
factorizes to: \vspace{-2mm}
\begin{align} 
\hat{y}^{MLE} &= \underset{y\in\{1,\dots,\vert\mathcal{C}\vert\}}{\mathrm{argmax}} \prod_{n=1}^N p(x_n\mid y)\\ 
                   &= \underset{y\in\{1,\dots,\vert\mathcal{C}\vert\}}{\mathrm{argmax}} \sum_{n=1}^N \log p(x_n\mid y)
\label{eq:multi-view-evidence-2}
\end{align} 
Thus the MLE of an object cluster $o$ belonging to one of the
$\mathcal{C}$ classes, is the class that corresponds to having the highest of
the sum of the log-likelihoods of their individual
class probabilities estimated for each of the $N$ observable viewpoints.










\section{Experiments}
\label{sec:experiments}
In this section, we evaluate the proposed SLAM-aware object
recognition method. In our experiments, we extensively evaluate our SLAM-aware
recognition system on the popular UW RGB-D
Dataset (v2)\cite{lai2011large,laiunsupervised}. We compare against the current
state-of-the-art solution proposed by~\citet{lai2012detection}, that utilize
full map and camera location information for improved
recognition performance. The UW RGB-D dataset contains a total 51 object
categories, however, in order to maintain a fair comparison, we consider the
same set of 5 objects as noted in~\cite{lai2012detection}. In experiment 3, we
propose scalable recognition solutions, increasing the number of objects
considered to all 51 object categories in the UW RGB-D Dataset.\vspace{1mm}

\textbf{Experiment 1: SLAM-Aware Object Recognition Performance Evaluation}
\label{experiment-1} We train and evaluate our system on the UW RGB-D Scene
Dataset~\cite{lai2011large,laiunsupervised}, providing mean-Average Precision
(mAP) estimates (see Table~\ref{tab:mAP-SLAM}) for the object recognition task
and compare against existing methods~\cite{lai2012detection}. We split our
experiments into two categories: \vspace{1mm} 


 \begin{figure*}[!t]
    \scriptsize
    {
      \setlength{\tabcolsep}{0.3em}
      \begin{tabular}{lccccccccc}
        \toprule
        \multirow{2}{*}{\textbf{Method}} & \multirow{2}{*}{\textbf{View(s)}} & \multirow{2}{*}{\textbf{Input}} & \multicolumn{7}{c}{\textbf{Precision/Recall}}\vspace{0.5mm}\\ 
        \multicolumn{2}{c}{} & & Bowl & Cap & Cereal Box & Coffee Mug & Soda Can &
                                                                                 Background & Overall\\\midrule
        DetOnly~\cite{lai2012detection} & Single 
                                         & RGB & 46.9/90.7 & 54.1/90.5 & 76.1/90.7 & 42.7/74.1 & 51.6/87.4 & 98.8/93.9 & 61.7/87.9\\
        \rowcolor{gray!25}
        Det3DMRF~\cite{lai2012detection} & Multiple 
                                         & RGB-D & 91.5/85.1 & 90.5/91.4 & 93.6/94.9 & 90.0/75.1 & 81.5/87.4 & 99.0/99.1 & \textbf{91.0}/88.8\\ 
        HMP2D+3D~\cite{laiunsupervised} & Multiple 
                                         & RGB-D & 97.0/89.1 & 82.7/99.0 & 96.2/99.3 & 81.0/92.6 & 97.7/98.0 & 95.8/95.0 & \textbf{90.9}/95.6\\ \midrule

        \rowcolor{gray!25}
        \textit{Ours} & Single  & RGB & 88.6/71.6 & 85.2/62.0 & 83.8/75.4 & 70.8/50.8 & 78.3/42.0 & 95.0/90.0 & \textbf{81.5}/59.4\\ 
        \textit{Ours} & Multiple & RGB & 88.7/70.2 & 99.4/72.0 & 95.6/84.3 & 80.1/64.1 & 89.1/75.6 & 96.6/96.8 & 89.8/72.0\\

        \bottomrule

      \end{tabular}
      \qquad
      \hspace{-4mm}
      \begin{tabular}{l}
        \includegraphics[width=0.5\columnwidth]{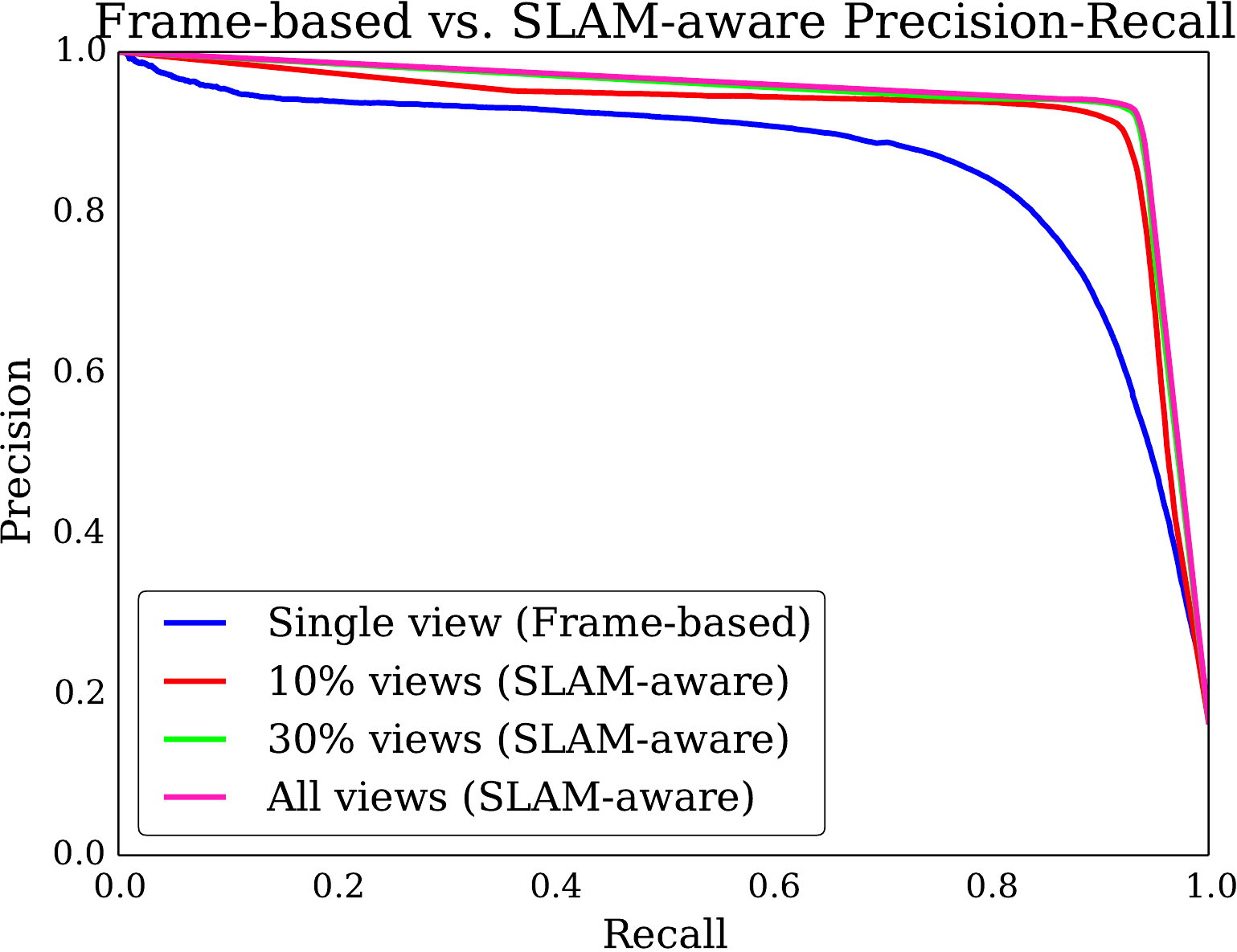}
      \end{tabular}
    }

    \captionlistentry[figure]{entry for figure}
    \addtocounter{figure}{-1}
    \label{fig:mAP-SLAM}
    \captionlistentry[table]{entry for table}
    \label{tab:mAP-SLAM}


    \captionsetup{labelformat=andfigure}
    \caption{\textbf{Left:} Object classification results using the UW RGB-D Scene
      Dataset~\cite{lai2011large,laiunsupervised}, providing mean-Average Precision
      (mAP) estimates for both Single-View, and Multi-View object recognition
      approaches. We compare against existing methods(\cite{lai2012detection,laiunsupervised}) that
      use RGB-D information instead of relying only on RGB images, in our
      case. Recognition for the single-view approach is done on a \textit{per-frame}
      basis, where prediction performance is averaged across all frames across all
      scenes. For the multi-view approach, recognition is done on a \textit{per-scene}
      basis, where prediction performance is averaged across all
      scenes. \textbf{Right:} Performance
      comparison via precision-recall for the Frame-based vs. SLAM-aware object
      recognition. As expected, the performance of our proposed SLAM-aware solution
      increases with more recognition evidence is aggregated across multiple viewpoints.\vspace{-2mm}
    }
  \end{figure*}

\textit{(i) Single-View recognition performance: } First, we evaluate the
recognition performance of our proposed system on each of the scenes in the
UW-RGB-D Scene Dataset on a per-frame basis, detecting and classifying objects
that occur every 5 frames in each scene (as done
in~\cite{lai2012detection}). Each object category is trained from images in the
Object Dataset, that includes several viewpoints of object instances with their
corresponding mask, and category information. Using training parameters
identical to the previous experiment, we achieve a performance of
81.5 mAP as compared to the detector performance of 61.7 mAP
reported in~\cite{lai2012detection}. Recognition is done on a per-image
basis, and averaged across all test images for reporting.
Figure~\ref{fig:frame-detection-results} shows the recognition results of our
system on a per-frame basis. We ignore regions labeled as background in the
figure for clarity and only report the correct and incorrect predictions in
green and red respectively. \vspace{1mm}

\begin{figure}[h]
  \centering
    \includegraphics[width=\columnwidth]{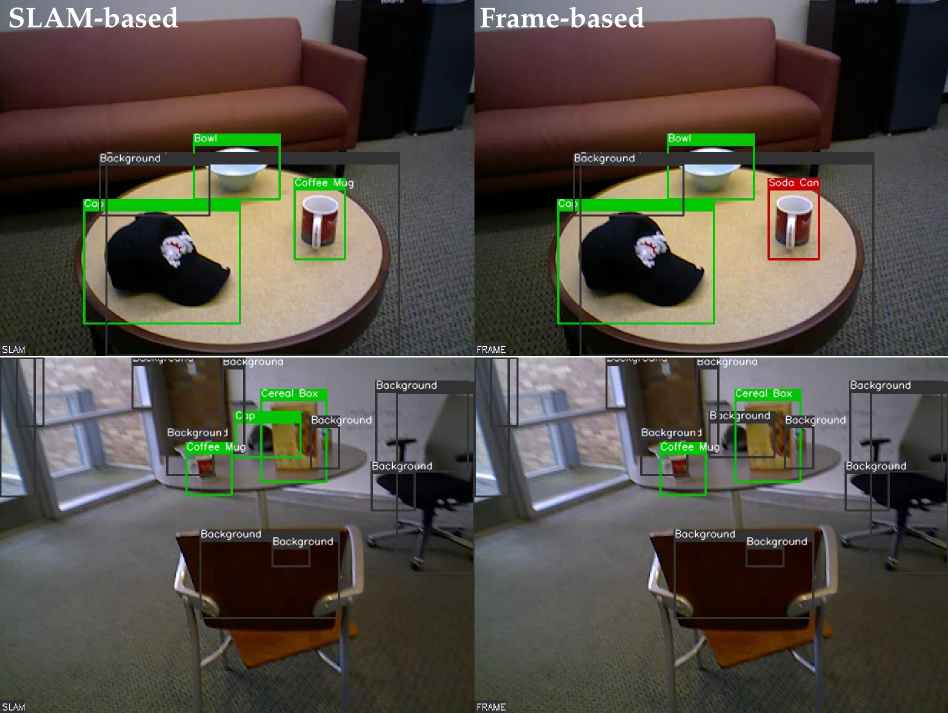} 
    \caption{More illustrations of the superior performance of the
        SLAM-aware object recognition in scenarios of ambiguity and
        occlusions. The coffee mug is misidentified as a soda can, and the cap
        in the bottom row is occluded by the cereal box.}
    \label{fig:more-slam-detect}
\end{figure}

\textit{(ii) Multi-View recognition performance: } In this section, we
investigate the performance of a SLAM-aware object recognition system. We
compare this to a SLAM-oblivious object detector described previously, and
evaluate using ground truth provided. Using the poses $\xi$ and reconstructed
map $\mathcal{M}$, multi-view object candidates are proposed and projected onto
each of the images for each scene sequence. Using the candidates provided as
input to the recognition system, the system predicts the likelihood and
corresponding category of an object (including background) contained in a
candidate bounding box. For each of the objects $o \in \mathcal{O}$ proposed,
the summed log-likelihood is computed (as in Eqn.~\ref{eq:multi-view-evidence})
to estimate the most likely object category over all the images for a particular
scene sequence. We achieve 89.8 mAP recognition performance on the 5 objects in
each of the scenes in~\cite{laiunsupervised} that was successfully reconstructed
by the ORB-SLAM-based semi-dense mapping
system. Figures~\ref{fig:intro-fig},~\ref{fig:multi-view-objectness},
~\ref{fig:slam-detection-results-1} and~\ref{fig:more-slam-detect} illustrate
the capabilities of the proposed system in providing robust recognition
performance by taking advantage of the monocular visual
SLAM-backend. Figure~\ref{fig:mAP-SLAM} illustrates the average precision-recall
performance on the UW RGB-D dataset, comparing the classical frame-based and our
SLAM-aware approach. As expected, with additional object viewpoints, our
proposed SLAM-aware solution predicts with improved precision and recall. In
comparison to that of HMP2D+3D~\cite{laiunsupervised}, they achieve only
slightly higher overall recognition performance of 90.9 mAP, as their
recognition pipeline takes advantage of the RGB and depth input to improve
overall scene reconstruction. We do note that while we perform comparably with
HMP2D+3D~\cite{laiunsupervised}, our BoVW+FLAIR architecture allows our system
to scale to a large number of object categories with \textit{near-constant
run-time}. We investigate the run-time performance and scalability concerns
further in Experiment 3.  \vspace{1mm}

%

%
%


\begin{figure}[b]
  \centering
  \vspace{2mm}
    \includegraphics[width=0.95\columnwidth]{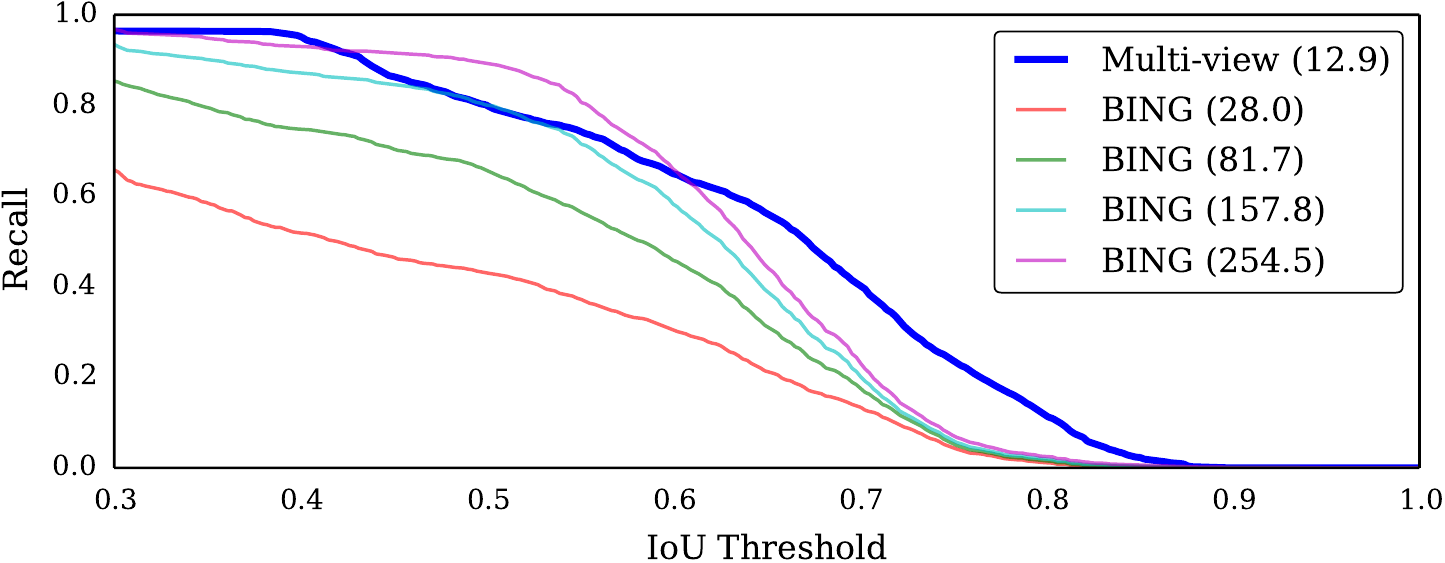} 
    \caption{\textit{Varying number of proposals:} We experiment with varied
      number of bounding boxes for the BING object proposal method, and compare
      against our multi-view object proposal method that uses considerably fewer
      number of bounding boxes to get similar or better recall rates. The numbers next
      to the label indicate the average number of windows proposed in the image. }
    \label{fig:objectness-performance}
\end{figure}

\textbf{Experiment 2: Multi-View Objectness}
\label{subsec:slam-driven-objectness} In this experiment, we investigate the
effectiveness of our multi-view object proposal method in identifying
category-independent objects in a continuous video stream. We compare the recall
of our object proposal method with the recently introduced
BING~\cite{cheng2014bing} object proposal technique, whose performance in
detection rate (DR) and run-time claim to be promising. We compare against the
BING method, varying the number of proposed object candidates by picking
proposals in descending order of their objectness
score. Figure~\ref{fig:objectness-performance} compares the overall performance
of our multi-view object proposal method that achieves better recall
rates, for a particular IoU threshold with
considerably fewer object proposals. The results provided are evaluated
on all the scenes provided in the UW-RGB-D dataset
(v2)~\cite{laiunsupervised}. 

\textbf{Experiment 3: Scalable recognition and run-time evaluation}
\label{subsec:results-run-time-perf} In this section, we investigate the
run-time performance of computing VLAD with integral histograms (FLAIR) for our
system and compare against previously proposed
approaches~\cite{van2014fisher,lai2012detection}. We measure the average speed
for feature-extraction (Dense-SIFT) and feature-encoding (VLAD) as they take up
over 95\% of the overall compute time. All experiments were conducted with a
single-thread on an Intel Core-i7-3920XM (2.9GHz). 




\begin{table}[h]
\centering
\scriptsize
\rowcolors{2}{gray!25}{white}
\begin{tabular}{lccc}
\toprule
\textbf{Method} & \textbf{$\vert\mathcal{C}\vert$} & \textbf{Run-time (s)} & \textbf{mAP/Recall} \\ \midrule
DetOnly~\cite{lai2012detection}   &  5 & ~\tiny$\approx$\scriptsize~1.8 s & 61.7/87.9\\ 
DetOnly~\cite{lai2012detection}   &  51 & ~$\ge5^{\dagger}$~s & -\\ 
HMP2D+3D~\cite{laiunsupervised}    &  9 & ~\tiny$\approx$\scriptsize~4~~ s & 92.8/95.3\\ \midrule
\textit{Ours}  & 5 & \textbf{1.6}~s & 81.5/59.4\\ %
\textit{Ours}  & 10 & \textbf{1.6}~s & 86.1/58.4\\ %
\textit{Ours}  & 51 & \textbf{1.7}~s & 75.7/60.9\\ \bottomrule %
\end{tabular}\\\vspace{1mm}
${}^{\dagger}${\tiny Expected run-time for sliding-window approaches as used in~\cite{lai2012detection}.}
\caption{Analysis of run-time performance of our system (for \textbf{frame-based
  detection}) compared to that
of~\cite{lai2012detection} and~\cite{laiunsupervised}. We achieve comparable
performance, and show scalable recognition performance with a near-constant run-time
cost (with increasing number of identifiable object categories
$\vert\mathcal{C}\vert = 51$). Existing sliding-window approaches become
impractical ($\ge4$ s run-time) in cases where $\vert\mathcal{C}\vert \approx
51$.}
\label{table:runtime}
\end{table}

\citet{van2014fisher} reports that the overall feature extraction
and encoding takes 5.15s (VQ 0.55s, FLAIR construction 0.6s, VLAD+FLAIR 4.0s)
per image, with the following parameters (2px step size, 3 Pyr. Scales,
[1x1],~[4x4] spatial pyramid bins). With significantly fewer candidate
proposals, and careful implementation, our system is able to achieve the same
(with 4px step size) in
approximately 1.6s. With reference to~\cite{lai2012detection}, where the
run-time performance of the sliding-window approach is directly proportional to
the number of object categories detectable, the authors report an overall
run-time of 1.8s for 5 object categories. However, scaling up their detection to
larger number of objects would imply costly runtimes, making it highly
impractical for real-time purposes. The run-time of our approach (based
on~\cite{van2014fisher}), on the other hand, is scalable to a larger number of
object categories, making it a strong contender for real-time recognition
systems. We summarize the run-times of our approach compared to that
of~\cite{lai2012detection} and~\cite{laiunsupervised} in Table~\ref{table:runtime}. \vspace{1mm}

\textbf{Discussion and Future Work} While there are benefits to running a
monocular visual SLAM-backend for recognition purposes, the inter-dependence of
the recognition system on this backend makes it vulnerable to the same
robustness concerns that pertain to monocular visual SLAM. In our experiments,
we noticed inadequacies in the semi-dense vSLAM implementation that failed to
reconstruct the scene on few occasions. To further emphasize recognition
scalability, we are actively collecting a larger scaled dataset (in increased
map area, and number of objects) to show the extent of capabilities of the
proposed system. Furthermore, we realize the importance of real-time
capabilities of such recognition systems, and intend to generalize the
architecture to a streaming approach in the near future. We also hope to release
the source code for our proposed method, allowing scalable and customizable
training with fast run-time performance during live operation.



\section{Conclusion} In this work, we develop a SLAM-aware object-recognition
system, that is able to provide robust and scalable recognition performance as
compared to classical SLAM-oblivious recognition methods. We leverage some of
the recent advancements in semi-dense monocular SLAM to propose objects in the
environment, and incorporate efficient feature encoding techniques to provide an
improved object recognition solution whose run-time is \textit{nearly-constant}
to the number of objects identifiable by the system. Through various
evaluations, we show that our SLAM-aware monocular recognition solution is
competitive to current state-of-the-art in the {RGB-D} object recognition
literature. We believe that robots equipped with such a monocular system will be
able to robustly recognize and accordingly act on objects in their environment,
in spite of object clutter and recognition ambiguity inherent from certain
object viewpoint angles.
~

\section*{Acknowledgments} This work was funded by the Office of Naval Research
under grants MURI N00014-10-1-0936, N00014-11-1-0688 and N00014-13-1-0588 and by
the National Science Foundation under Award IIS-1318392. We would like to thank
the authors of ORB-SLAM and LSD-SLAM for providing source code of their work,
and the authors of the UW-RGB-D Dataset~\cite{lai2012detection,laiunsupervised}
for their considerable efforts in collecting, annotating and developing
benchmarks for the dataset. \pagebreak


\bibliographystyle{abbrvnat_ordered} 
{\small
\bibliography{tex/references}
}

\end{document}